\documentclass[accepted]{article}
\usepackage{aistats2025}
\usepackage{natbib}
\setcitestyle{open=(,close=)}
\usepackage{graphicx}
\usepackage{amsmath}
\usepackage{algorithmic}
\usepackage{mathtools}
\usepackage{booktabs}

\usepackage{amsmath,amsfonts,bm}

\def\eqref#1{equation~\ref{#1}}

\def\1{\bm{1}}

\DeclareMathAlphabet{\mathsfit}{\encodingdefault}{\sfdefault}{m}{sl}
\SetMathAlphabet{\mathsfit}{bold}{\encodingdefault}{\sfdefault}{bx}{n}

\usepackage{hyperref}
\usepackage{url}
\usepackage{caption}
\usepackage{subcaption}
\usepackage{xr}
\usepackage{svg}
\usepackage[ruled,linesnumbered]{algorithm2e}
\usepackage{amsthm}
\usepackage{amssymb}
\newtheorem{theorem}{Theorem}
\RestyleAlgo{ruled}
\usepackage{amsfonts}
\usepackage{multicol}
\usepackage{wrapfig}

\usepackage{lmodern}
\usepackage{anyfontsize}

\begin{document}
\twocolumn[
\aistatstitle{ Parallel Backpropagation for Inverse of a Convolution with Application to Normalizing Flows } 
\aistatsauthor{ Sandeep Nagar \And Girish Varma }
\aistatsaddress{Machine Learning Lab, IIIT Hyderabad, India\\
\url{https://naagar.github.io/InverseFlow/}}]

\begin{abstract}
The inverse of an invertible convolution is an important operation that comes up in Normalizing Flows, Image Deblurring, etc. The naive algorithm for backpropagation of this operation using Gaussian elimination has running time $O(n^3)$ where $n$ is the number of pixels in the image. We give a fast parallel backpropagation algorithm with running time $O(\sqrt{n})$ for a square image and provide a GPU implementation of the same. Inverse of Convolutions are usually used in Normalizing Flows in the sampling pass, making them slow. We propose to use the Inverse of Convolutions in the forward (image to latent vector) pass of the Normalizing flow. Since the sampling pass is the inverse of the forward pass, it will use convolutions only, resulting in efficient sampling times. We use our parallel backpropagation algorithm to optimize the inverse of the convolution layer, resulting in fast training times. We implement this approach in various Normalizing Flow backbones, resulting in our Inverse-Flow models. We benchmark Inverse-Flow on standard datasets and show significantly improved sampling times with similar bits per dimension compared to previous models. 
\end{abstract}


\section{Introduction}
Large-scale neural network optimization using gradient descent is made possible due to efficient and parallel back-propagation algorithms \citep{bottou2010large}. Large models could not be trained on large datasets without such fast back-propagation algorithms. All operations for building practical neural network models need efficient back-propagation algorithms \citep{lecun2002efficient}. This has limited types of operations that can be used to build neural networks. Hence, it is important to design fast parallel backpropagation algorithms for novel operations that could make models more efficient and expressive.

Convolutional layers are commonly used in Deep Neural Network models as they have fast parallel forward and backward pass algorithms \citep{lecun2002efficient}. The inverse of a convolution is a closely related operation with use cases in Normalizing Flows \citep{karami2019invertible}, Image Deblurring \citep{eboli2020end}, Sparse Blind Deconvolutions \citep{xu2014deep}, Segmentation, etc. However, the Inverse of a Convolution is not used directly as a layer for these problems since straightforward algorithms for the backpropagation of such layers are highly inefficient. Such algorithms involve computing the inverse of a very large dimensional matrix. 

Fast sampling is crucial for Normalizing flow models in various generative tasks due to its impact on practical applicability and real-time performance \citep{papamakarios2021normalizing}. Rapidly producing high-quality samples is essential for large-scale data generation and efficient model evaluation in fields such as image generation, molecular design \citep{zang2020moflow}, image deblurring, and deconvolution. Normalizing flows have demonstrated their capability in constructing high-quality images \citep{kingma2018glow, meng2022butterflyflow}. However, the training and sampling process is computationally expensive due to the repeated need for inverting functions (e.g., convolutions) \citep{lipman2023flow}. Existing approaches rely on highly constrained architectures and often impose limitations like diagonal, triangular, or low-rank Jacobian matrices and approximate inversion methods \citep{hoogeboom2019emerging, keller2021self}. These constraints restrict the expressiveness and efficiency of normalizing flow models. To overcome these limitations, fast, efficient, and parallelizable algorithms are needed to compute the inverse of convolutions and their backpropagation, along with GPU-optimized implementations. Addressing these challenges would significantly enhance the performance and scalability of Normalizing flow models.

In this work, we propose a fast, efficient, and parallelized backpropagation algorithm for the inverse of convolution with running time $O(mk^2)$ on an $m\times m$ input image. We provide a parallel GPU implementation of the proposed algorithm  (together with baselines and experiments) in CUDA. Furthermore, we design \emph{Inverse-Flow}, using an inverse of convolution ($f^{-1}$) in forward pass and convolution ($f$) for sampling. Inverse-flow models generate faster samples than standard Normalizing flow models.  

In summary, our contribution includes:
\begin{enumerate}
    \item We designed a fast and parallelized backpropagation algorithm for the inverse of convolution operation.
    \item \sloppy Implementation of proposed backpropagation algorithm for inverse of convolution on CUDA-GPU. \fussy
    \item We propose a multi-scale flow architecture, \emph{Inverse-Flow}, for fast training of inverse of convolution using our efficient backpropagation algorithm and faster sampling with $k \times k$ convolution.
    
    \item Benchmarking of \emph{Inverse-Flow} and a small linear, 9-layer flow model on MNIST, CIFAR10 datasets.
    
\end{enumerate}

\begin{figure}
    \centering
    \includegraphics[width=0.99\linewidth]{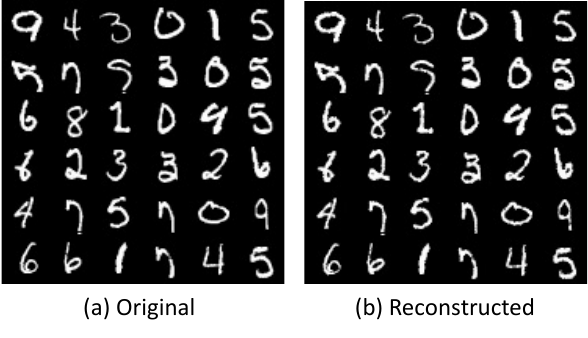}
    \label{fig:recon_img}
    \caption{a). Images from MNIST dataset. b). Reconstructed images using an Inverse-Flow model based on the inv-conv layer for a forward pass.}
\end{figure}
\section{Related work}

\paragraph{Backpropagation for Inverse of Convolution}
The backpropagation algorithm performs stochastic gradient descent and effectively trains a feed-forward neural network to approximate a given continuous function over a compact domain. \citep{hoogeboom2019emerging} proposed invertible convolution, Emerging, generalizing 1x1 convolution from Glow \citep{kingma2018glow}. \citep{finzi2019invertible} proposed periodic convolution with $k \times k$ kernels. Emerging convolution combines two autoregressive convolutions \citep{kingma2016improved}, and parallelization is not possible for its inverse. MaCow \citep{ma2019macow} uses four masked convolutions in an autoregressive fashion to get a receptive field of $3 \times 3$ standard convolution, which leads to slow sampling and training. To our knowledge, this work is the first to propose a backpropagation algorithm for the inverse of convolution. Additionally, it is the first to utilize an inverse Normalizing flow for training and a standard flow for sampling, marking a novel approach in the field.

\paragraph{Normalizing flows (NF)}
NF traditionally relies on invertible specialized architectures with manageable Jacobian determinants \citep{keller2021self}. One body of work builds invertible architectures by concatenating simple layers (coupling blocks), which are easy to invert and have a triangular Jacobian \cite{nagar2021cinc}. Many choices for coupling blocks have been proposed, such as MAF \citep{papamakarios2017masked}, RealNVP \citep{dinh2016density}, Glow \citep{kingma2018glow}, Neural Spline Flows \citep{durkan2019neural}. Self Normalizing Flow (SNF) \citep{keller2021self} is a flexible framework for training NF by replacing expensive terms in gradient by learning approximate inverses at each layer. Several types of invertible convolution emerged to enhance the expressiveness of NF. Glow has stood out for its simplicity and effectiveness in density estimation and high-fidelity synthesis.

\paragraph{Autoregressive}
\citep{kingma2016improved} propose an inverse autoregressive flow and scale well to high dimensions latent space, which is slow because of its autoregressive nature. \cite{papamakarios2017masked} introduced NF for density estimation with masked autoregressive. Sample generation from autoregressive flows is inefficient since the inverse must be computed by sequentially traversing through autoregressive order \citep{ma2019macow}

\paragraph{Invertible Neural Network}
\citep{dinh2016density} proposed Real-NVP, which uses a restricted set of non-volume preserving but invertible transformations. \citep{kingma2018glow} proposed Glow, which generalizes channel permutation in Real-NVP with $1 \times 1$ convolution. 
However, these NF-based generative models resulted in worse sample generation compared to state-of-the-art autoregressive models and are incapable of realistic synthesis of large images compared to GANs \citep{brock2018large} and Diffusion Models. CInC Flow \citep{nagar2021cinc} proposed a fast convolution layer for NF. ButterflyFlow \citep{meng2022butterflyflow} leverage butterfly layers for NF models. FInC Flow \citep{visapp23} leverages the advantage of parallel computation for the inverse of convolution and proposed efficient parallelized operations for finding the inverse of convolution layers and achieving $O(n\times k^2)$. We designed a backpropagation algorithm for the inverse of convolution layers. Then,  multi-scale architecture, Inverse-Flow, is designed using an inverse of convolution for forward pass and convolution for sampling pass and backward pass.

\paragraph{Sampling Time} NF requires large and deep architectures to approximate complex target distributions \citep{cornish2020relaxing} with arbitrary precision. \cite{jung2024normalizing} present the importance of fast sampling for NF models. Modeling distribution using NF models requires the inverse of a series of functions, a backward pass, which is slow. This creates a limitation of slow sample generation. To address this, we propose Inverse-Flow, which uses convolution (fast parallel operation, $O(k^2)$, $k\times k$= kernel size) for a backward pass and inverse of convolution for a forward pass.



\section{Fast Parallel Backpropagation for Inverse of a Convolution.}

\begin{figure}
    \centering
    \includegraphics[width=0.99\linewidth]{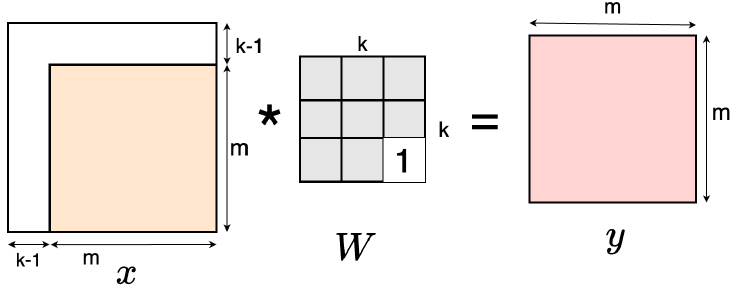}
    \caption{Invertible convolution with zero padding (top, left) on input $x$ and masking of kernel $W_{k,k} = 1$}
    \label{fig:inv_conv}
\end{figure}

We assume that the input/output of a convolution is an $m \times m$ image, with the channel dimension assumed to be 1 for simplicity. The algorithm can naturally be extended to any number of channels. We also assume that input to convolution is padded on top and left sides with $k-1$ zeros, where $k$ is a kernel size; see Figure \ref{fig:inv_conv}. Furthermore, we assume that the bottom right entry of the convolution kernel is 1, which ensures that it is invertible. More details about assumptions are in section \ref{sec:pad}.

The convolution operation is a Linear Operator (in Linear Algebraic terms; see Figure \ref{fig:inv_conv}) on space of $m \times m$ matrices. Considering this space as column vectors of dimension $m^2$, this operation corresponds to multiplication by a $m^2 \times m^2$ dimensional matrix. Hence inverse of convolution is also a linear operator represented by a $m^2 \times m^2$ dimensional matrix. Suppose vectorization of $m \times m$ matrix to $m^2$ is done by row-major ordering; diagonal entries of Linear Operator matrix will be the bottom right entry of kernel, which we have assumed to be 1.

While convolution operation has fast parallel forward and backpropagation algorithms with running time $O(k^2)$ (assuming there are $O(m^2)$ parallel processors), a naive approach for the inverse of convolution using Gaussian Elimination requires $O(m^6)$. \citep{visapp23} gave a fast parallel algorithm for the inverse of convolution with running time $O(mk^2)$. In this section, we give a fast parallel algorithm for backpropagation of the inverse of convolution (inv-conv) with running time $O(mk^2)$ (see Table \ref{tab:run_time}). Our backpropagation algorithm allows for efficient optimization of the inverse of convolution layers using gradient descent.

\begin{table}[!ht]
    \centering
    \caption{Running times of algorithms for Forward and Backward passes assuming there are enough parallel processors as needed. The forward pass algorithm for the Inverse of Convolution was improved by \citep{visapp23}. In this work, we give an efficient backward pass algorithm for Inverse of Convolution.}
    \begin{tabular}{lrr}
    \toprule
       Layer & Forward & Backpropagation \\ 
    \midrule
    Std. Conv. & $O(k^2)$ & $O(k^2)$ \\
    inv-conv (naive)& $O((m^2)^{3})$ & $O((m^2)^{3})$ \\
    inv-conv. & $ O(mk^2)$  & $\bf O(mk^2)$ \\
    \bottomrule
    \end{tabular}
    \label{tab:run_time}
\end{table}

\paragraph{Notation:} We will denote input to the inverse of convolution (inv-conv) by $y \in \mathbb R^{m^2}$ and output to be $x \in \mathbb R^{m^2}$. We will be indexing $x,y$ using $p = (p_1, p_2) \in \{ 1, \cdots, n\} \times \{ 1, \cdots, n\}$. We define 
$$\Delta(p) = \{ (i,j):  0 \leq p_1 - i, p_2 -j < k  \} \setminus \{ p \}.$$
$\Delta(p)$ informally is a set of all pixels except $p$ in the input, which depends on $p$ in the output when convolution is applied with top and left padding. We also define a partial ordering $\leq$ on pixels as follows
$$ p \leq q \quad \Leftrightarrow \quad p_1 \leq q_1 \text{ and } p_2 \leq q_2.$$

The $k\times k$ convolution kernel is given by matrix $W \in \mathbb R^{k \times k}$. For the backpropagation algorithm for inv-conv, the input is 
$$x \in \mathbb R^{m^2} \text{ and } \frac{\partial L}{\partial x} \in \mathbb R^{m^2},$$ 
where $L$ is the loss function. We can compute $y$ on $O(m^2k^2)$ time using the parallel forward pass algorithm of \citep{visapp23}. The output of backpropagation algorithm is $$\frac{\partial L}{\partial y} \in \mathbb R^{m^2} \text{ and } \frac{\partial L}{\partial W} \in \mathbb R^{k^2}$$
which we call input and weight gradient, respectively. We provide the algorithm for computing these in the next 2 subsections.

\subsection{Computing Input Gradients}
Since $y$ is input to inv-conv and $x$ is output, $y = \text{conv}_W(x)$ and we get the following $m^2$ equations by definition of the convolution operation. 
\begin{equation} \label{eqn:conv}
  y_p = x_p + \sum_{q \in \Delta(p)} W_{(k,k) - p + q} \cdot  x_q 
\end{equation}
Using the chain rule of differentiation, we get that
\begin{equation}
    \frac{\partial L}{ \partial y_p} = \sum_{q} \frac{\partial L}{ \partial x_q} \times \frac{\partial x_q}{ \partial y_p}.
\end{equation}
Hence if we find $\frac{\partial x_q}{ \partial y_p}$ for every pixels $p,q$, we can compute $ \frac{\partial L}{ \partial y_p}$ for every pixel $p$.

\begin{theorem}\label{the:dx_dy}
$$
\frac{\partial x_q}{ \partial y_p} = 
\begin{cases}
\quad    1 - \sum_{q \in \Delta(p)} W_{(k,k) - p + q} \cdot  \frac{\partial x_q}{\partial y_p}  & \text{ if } p = q\\
   \quad 0 & \text{ if } q \not \leq p\\
    - \sum_{r \in \Delta(p)} W_{(k,k) - r} \frac{\partial x_{p-r'}}{\partial y_p} & \text{ otherwise. }
\end{cases}
$$
\end{theorem}
Formal proofs are deferred to the supplementary.
\paragraph{Informal proof:}

The theorem presents computing $\frac{\partial x_q}{\partial y_p}$, which represents how a change in input pixel $y_p$ affects output pixel $x_q$ in an inverse of the convolution operation. Let's break down each case:

\emph{Case 1:} When $p = q$, take partial derivative with respect to $y_p$ on both sides of Equation \ref{eqn:conv} and rearranging. 
$$\frac{\partial x_p}{\partial y_p} = 1 - \sum_{q \in \Delta(p)} W_{(k,k) - p + q} \cdot  \frac{\partial x_q}{\partial y_p}  $$

So if $\frac{\partial x_q}{\partial y_p}$ is known for all $q \leq p$, we can compute $\frac{\partial x_p}{\partial y_p}.$ Since the off-diagonal entries are unrelated in the $\leq$ partial order, we can compute all of them in parallel, provided the previous off-diagonal entries are known.

\emph{Case 2:} From Equation \ref{eqn:conv}, when $q \not\leq p$, we have: $\frac{\partial x_q}{\partial y_p} = 0$. This case uses partial ordering defined earlier. If $q$ is not less than or equal to $p$ in this ordering, it means that output pixel $x_q$ is not influenced by input pixel $y_p$ in the inverse of convolution operation \ref{eqn:conv}. Therefore, the derivative is 0.

\emph{Case 3:} For all other cases:

$$\frac{\partial x_q}{\partial y_p} = -\sum_{r \in \Delta(p)} W_{(k,k) - r} \frac{\partial x_{p-r'}}{\partial y_p}$$

\begin{itemize}
    \item $\Delta(p)$ is set of all pixels (except $p$) that depend on $p$ in a regular convolution operation.
    \item $W_{(k,k) - r}$ represents weight in convolution kernel corresponding to relative position of $r$.
    \item $\frac{\partial x_{p-r'}}{\partial y_p}$ is a recursive term, representing how changes in $y_p$ affect $x$ at a different position.
\end{itemize}

The negative sign and summation in this formula account for the inverse nature of the operation and cumulative effects of the convolution kernel.

\subsection{Computing Weight Gradients}
From Equation \ref{eqn:conv}, we can say computing the gradient of loss $L$ with respect to weights $W$ involves two key factors. Direct influence: how a specific weight $W_{a}$ in convolution kernel directly affects output $x$ pixels, and Recursive Influence: how neighboring pixels, weighted by the kernel, indirectly influence output $x$ during convolution operation. Similarly, to compute  gradient of  loss $L$ w.r.t filter weights \( W \), we apply  chain rule:

\begin{equation} \label{eq:dl_dw}
    \frac{\partial L}{\partial W} = \frac{\partial L}{ \partial x} * \frac{\partial x}{\partial W}
\end{equation}

where: \( \frac{\partial L}{\partial x} \) is  gradient of  loss with respect to  output \( x \) and convolution operation is applied between \( \frac{\partial L}{\partial x} \) and  output \( x \). Computing the gradient of loss $L$ with respect to convolution filter weights $W$ is important in backpropagation when updating the convolution kernel during training. Similarly, $\partial L/ \partial W$ can be calculated as \ref{eq:dl_dw} and $\partial x/\partial W$ can be calculated as (\ref{eq:dx_dw}) for each $k_{i, j}$ parameter by differentiating \ref{eqn:conv} w.r.t $W$:

\begin{equation}
    \label{eq-dw}
    \frac{\partial L}{ \partial W_a} =  \sum {\frac{\partial L}{\partial x_q} * \frac{\partial x_q}{\partial W_a}}
\end{equation}

Equation \ref{eq-dw} states that to compute the gradient of loss with respect to each weight $W_a$, we need to:
\begin{itemize}
    \item Compute how loss $L$ changes with respect to each output pixel $x_q$ (denoted by $\frac{\partial L}{ \partial x_q}$).
    \item Multiply this by gradient of each output pixel $x_q$ with respect to weight $W_a$ (denoted by $\frac{\partial x_q}{ \partial W_a}$)
\end{itemize}

We then sum over all output pixels $x_q$.

\begin{theorem} \label{eq:dx_dw}
$$
\frac{\partial x_q}{\partial W_a} = 
\begin{cases}
    0 & \text{if } q \leq a \\
    -\sum_{q' \in \Delta q(a)} W_{q' - a} \cdot \frac{\partial x_{q-q'}}{\partial W_a} - x_{q-a} & \text{if } q > a
\end{cases}
$$
\end{theorem}
Formal proofs are deferred to the supplementary.
\paragraph{Informal proof:} Computation of $\partial x_q/\partial W_a$ depends on relative positions of pixel $q$ and kernel weight index. 

\emph{Case 1:} When $a \leq q $, if index of weight matches index of output pixel \ref{eqn:conv}, gradient is 0. This means that weight does not directly influence the corresponding pixel in this case. \\
\emph{Case 2:} When $ q > a$,  gradient is computed recursively by summing over neighboring pixel positions $q'$ in convolution window. In this case, $q' \in \Delta q(a)$ represents pixels within the kernel's influence around pixel $q$ that specifically correspond to weight $W_a$, meaning pixels whose relative position to $q$ makes them affected by the particular weight $W_a$ during the convolution operation. The convolution kernel weights $W_{q'-a}$ and shifted pixels value $x_{q-a}$ are used to calculate gradients. See the Supplement section for more elaborated proof.

\subsection{Backpropagation Algorithm for Inverse of Convolution}

The backpropagation algorithm for the inverse of convolution (inv-conv) computes gradients necessary for training models that use inv-conv operation for a forward pass. Our proposed algorithm \ref{algo:bp_cuda} efficiently calculates gradients with respect to both input ( $\frac{\partial L}{\partial Y}$) and convolution kernel ($\frac{\partial L}{\partial K}$) using a parallelized GPU approach.

Given the gradient of loss $L$ with respect to output ($\frac{\partial L}{\partial X}$ ), the algorithm updates input gradient $\frac{\partial L}{\partial Y}$ by accumulating contributions from each pixel in output, weighted by corresponding kernel values. Simultaneously,  kernel gradient $\frac{\partial L}{\partial K}$ is computed by accumulating contributions from spatial interactions between input and output. The process is parallelized across multiple threads, with each thread handling updates for different spatial and channel indices, ensuring efficient execution. This approach ensures that both input and kernel gradients are computed in a time-efficient manner, making it scalable for high-dimensional inputs and large kernels. A fast algorithm is key for enabling gradient-based optimization in models involving the inverse of convolution.

\paragraph{Complexity of Algorithm \ref{algo:bp_cuda}:} This computes $\frac{\partial L}{ \partial y}$ and $\frac{\partial L}{ \partial w}$ in $O(mk^2)$ utilizing independence of each diagonal of output $x$ and sequencing of $m$ diagonals. Diagonals are processed sequentially, but elements within each diagonal are processed in parallel. Each diagonal computation takes $O(k^2)$ time due to $k \times k$ kernel. This results in a time complexity of total $O(mk^2)$ and represents a substantial improvement over the naive $O(m^6)$ approach. It makes algorithm highly efficient and practical for use in deep learning models with inverse of convolution layers, even for large input sizes or kernel sizes.

\SetKwComment{Comment}{/* }{ */}

\begin{algorithm}[!ht]\caption{Backpropagation Algorithm for Inverse of Convolution (Input and Weight Gradients)} \label{algo:bp_cuda} 
\KwIn{$K$: Kernel of shape $(C, C, k_H, k_W)$\\ $Y$: output of conv of shape $(C, H, W)$\\ 
$\frac{\partial L}{\partial X}$: gradient of shape $(C, H, W)$} 
\KwOut{$\frac{\partial L}{\partial Y}$: gradient of shape $(C, H, W)$\\ $\frac{\partial L}{\partial K}$: gradient of shape $(C, C, k_H, k_W)$}
\textbf{Initialization:} 

$\frac{\partial L}{\partial Y} \gets 0$ (initialize input gradient to zero)\\
$\frac{\partial L}{\partial K} \gets 0$ (initialize kernel gradient to zero)

\For{$d \gets 0, H + W - 1$}{
    \For{$c \gets 0, C - 1$}{
        \Comment{The below lines of code are executed parallelly on different threads on GPU for every index $(c, h, w)$ on $d$th diagonal.}
        \For{$k_h \gets 0, k_H - 1$}{
        \For{$k_w \gets 0, k_W - 1$}{
        \For{$k_c \gets 0, C - 1$}{
            
            \If{pixel $(k_c, h - k_h, w - k_w)$ not out of bounds}{
                \Comment{Compute input gradient for every pixel $(c, h, w)$:}
                
                $\frac{\partial L}{\partial Y}[c, h, w] \gets \frac{\partial L}{\partial Y}[c, h, w] + \frac{\partial L}{\partial X}[c, h, w] \cdot K[c, k_c, k_H - k_h - 1, k_W - k_w - 1]$

                \Comment{Compute kernel gradient:}
                
                $\frac{\partial L}{\partial K}[c, k_c, k_h, k_w] \gets \frac{\partial L}{\partial K}[c, k_c, k_h, k_w] + \frac{\partial L}{\partial X}[c, h, w] \cdot X[k_c, h - k_h, w - k_w]$
            }
        }
        }
        }
        \Comment{synchronize all threads}
    }
}
\Return{$\frac{\partial L}{\partial Y}, \frac{\partial L}{\partial K}$}
\end{algorithm}


            

\section{Normalizing Flows}

Normalizing flows are generative models that enable exact likelihood evaluation. They achieve this by transforming a base distribution into a target distribution using a series of invertible functions.

Let $\mathbf{z} \in \mathcal{Z}$ be a random variable with a simple base distribution $p_Z(\mathbf{z})$ (e.g., a standard Gaussian). A Normalizing flow transforms $\mathbf{z}$ into a random variable $\mathbf{y} \in \mathcal{Y} $ with a more complex distribution $p_Y(\mathbf{y})$ through a series of invertible transformations: $\mathbf{y} = f(\mathbf{z}) = f_1(f_2(\cdots f_K(\mathbf{z})))$. Probability density of transformed variable $\mathbf{y}$ can be computed using change-of-variables formula:
\begin{equation}\label{eq:flow}
p_Y(\mathbf{y}) = p_Z(\mathbf{z}) \left|\det\frac{\partial f^{-1}}{\partial \mathbf{y}}\right| = p_Z(f^{-1}(\mathbf{y})) \left|\det\frac{\partial f}{\partial \mathbf{z}}\right|^{-1},
\end{equation}

where $\left|\det\frac{\partial f}{\partial \mathbf{z}}\right|$ is absolute value of determinant of Jacobian of $f$.

This relationship (\ref{eq:flow}) can be modeled as $y = f_\theta(z)$ called change of variable formula, where $\theta$ is a set of learnable parameters. This formula enables us to compute the likelihood of $y$ as:
\begin{equation}
    \log p_Y(y) = \log p_Z(f_\theta(y)) + \log \left| \det \left( \frac{\partial f_\theta(y)}{\partial y} \right) \right|,
\end{equation}

where second term, $\log \left| \det \left( \frac{\partial f_\theta(y)}{\partial y} \right) \right|$, is log-determinant of Jacobian matrix of transformation $f_\theta$. This term ensures volume changes induced by transformation are properly accounted for in likelihood. For invertible convolutions, a popular choice for constructing flexible Normalizing flows, the complexity of computing the Jacobian determinant can be addressed by making it a triangular matrix with all diagonal entries as $1$, and the determination will always be one.

In this work, we leverage the fast inverse of convolutions for a forward pass (inv-conv = $f_{\theta}$) and convolution for a backward pass and designed the \emph{Inverse-Flow} model to generate fast samples. To train  Inverse-Flow, we use our proposed fast and efficient backpropagation algorithm for the inverse of convolution.
\begin{figure}[!ht]
    \centering
    \includegraphics[width=0.99\linewidth]{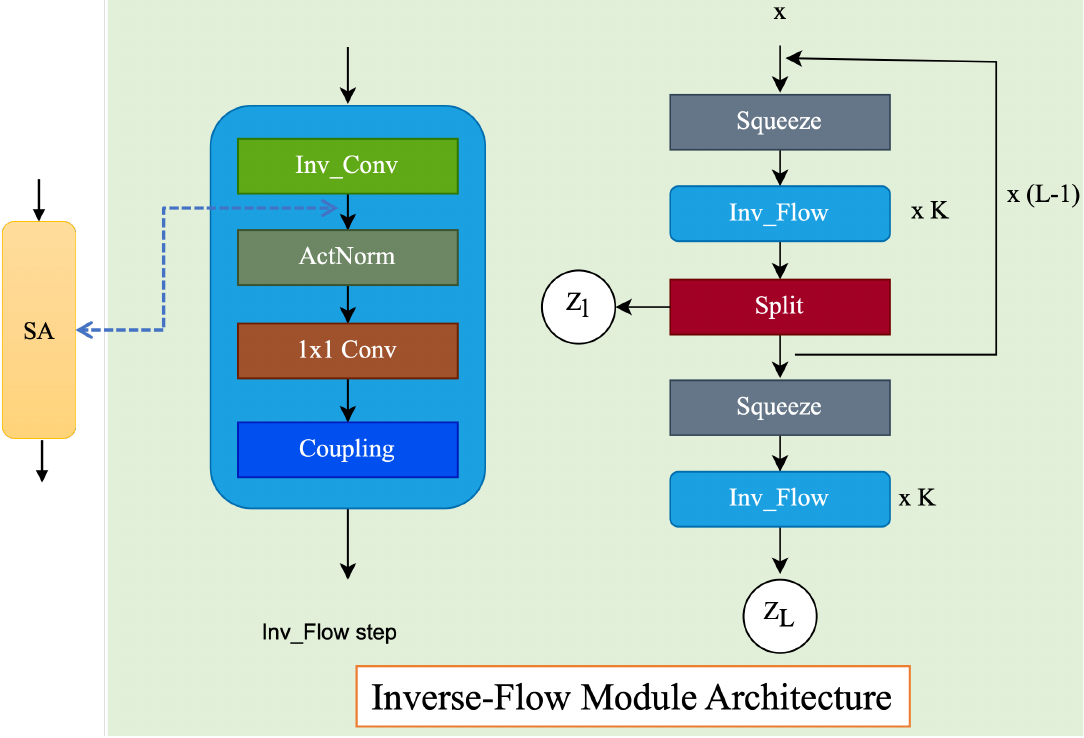}
    \caption{Multi-scale architecture of Inverse-Flow model and Inv Flow step.}
    \label{fig:ms_if_arch}
\end{figure}

\subsection{Inverse-Flow Architecture}
Figure \ref{fig:ms_if_arch} shows the architecture of Inverse-Flow. Designing flow architecture is crucial to obtaining a family of bijections whose Jacobian determinant is tractable and whose computation is efficient for forward and backward passes. Our model architecture resembles the architecture of Glow \citep{kingma2018glow}. Multi-scale architecture involves a block of Squeeze, an $Inv\_Flow$ Step repeated $K$ times, and a Split layer. The block is repeated $L - 1$ a number of times. A Squeeze layer follows this, and finally, the $Inv\_Flow$ Step is repeated $K$ times. At the end of each Split layer, half of the channels are ’split’ (taken away) and modeled as Gaussian distribution samples. These splits half channels are latent vectors. The same is done for output channels. These are denoted as $Z_{L}$ in Figure \ref{fig:ms_if_arch}. Each $Inv\_Flow$ Step consists of an $Inv-Conv$ layer, an Actnorm Layer, and a $1 \times  1$ Convolutional Layer, followed by a Coupling layer.

\paragraph{$Inv\_Flow$ Step:} First we consider inverse of convolution and call it Inv\_Conv layer. Figure \ref{fig:ms_if_arch} left visualizes the inverse of $k \times k$ convolution (Inv-Conv) block followed by Spline Activation layer.

\paragraph{SplineActivation (SA):} \cite{bohra2020learning} introduced a free-form trainable activation function for deep neural networks. We use this layer to optimize Inverse-Flow model. Figure \ref{fig:ms_if_arch}, left most: SA layer is added in Inv\_Flow step after Inv\_Conv block.

\paragraph{Actnorm:} Next, Actnorm, introduced in \citep{kingma2018glow}, acts as an activation normalization layer similar to that of a batch normalization layer. Introduced in Glow, this layer performs affine transformation of input using scale and bias parameters per channel.

\paragraph{ $1  \times 1$ Convolutional:} This layer introduced in Glow does a $1 \times 1$ convolution for a given input. Its log determinant and inverse are very easy to compute. It also improves the effectiveness of coupling layers.

\paragraph{Coupling Layer:} RealNVP \citep{dinh2016density} introduced a layer in which input is split into two halves. The first half remains unchanged, and the second half is transformed and parameterized by the first half. The output is the concatenation of the first half and affine transformation by functions parameterized by the first of the second half. The coupling layer consists of a $3 \times 3 $ convolution followed by a $1 \times 1$ and a modified $3 \times 3$ convolution used in Emerging.

\paragraph{Squeeze:} this layer takes features from spatial to channel dimension \citep{behrmann2019invertible}, i.e., it reduces feature dimension by a total of four, two across height dimension and two across width dimension, increasing channel dimension by four. As used by RealNVP, we use a squeeze layer to reshape feature maps to have smaller resolutions but more channels.

\paragraph{Split:} input is split into two halves across channel dimensions. This retains the first half, and a function parameterized by the first half transforms the second half. The transformed second half is modeled as Gaussian samples, which are latent vectors. We do not use the checkerboard pattern used in RealNVP and many others to keep the architecture simple.

\subsection{Inverse-Flow Training}
During training, we aim to learn the parameters of invertible transformations (including invertible convolutions) by maximizing the likelihood of data. Given input data $y$ and a simple base distribution $p_z$ (e.g., a standard Gaussian distribution), the training process aims to find a sequence of invertible transformations such that $z = \text{inv-conv}(y)$, where $z$ is a latent vector from base distribution and $\theta$ represents the model. The likelihood of data under the model is computed using the change of variables formula:
    $$\log_{p_Y}(y) = \log_{p_z}(\text{inv-conv}(y)) 
    + \log\Big|\det(\frac{\partial \text{inv-conv}(y)}{\partial y})\Big|$$
Here $\det(\frac{\partial \text{inv-conv}(y)}{\partial y})$ represents a Jacobian matrix of transformation, which is easy to compute for \emph{inv-conv}.
\subsection{Sampling for Inverse-flow}

To generate samples from the model after training, we use the reverse process: Sample from the base distribution $z \sim p_{z}(z)$ from a Gaussian distribution. Apply inverse of learned transformation to get back data space: 
$y =  \text{conv}_{\theta}(z) $
This process involves performing the inverse of all transformations in flow, including \emph{inv-conv}. This sampling procedure ensures that generated samples are drawn from the distribution that the model has learned during training, utilizing the invertible nature of convolutional layers.

\section{Results}
In this section, we compare the performance of Inverse-Flow against other flow architectures. We present Inverse-Flow model results for bit per-dimension ($\log$-likelihood), sampling time (ST), and forward pass time (FT) on two image datasets. To test the modeling of Inverse-Flow, we compare bits-per-dimension (BPD). To compare ST, we generate 100 samples for each flow setting on single \emph{NVIDIA GeForce RTX 2080 Ti GPU} and take an average of 5 runs after warm-up epochs. For comparing FT, we present forward pass time with a batch size of $100$, averaging over 10 batch runs after warm-up epochs. Due to computation constraints, we train all models for 100 epochs, compare BPD with other state-of-the-art, and show that Inverse-Flow outperforms based on model size and sampling speed. 

\begin{table}[ht]
    \centering
    \caption{Performance comparison  for MNIST dataset with $4$ block size and $2$ blocks, small model size. ST = samling time, FT = Forward pass, NLL is negative-$\log$-likelihood. Inverse-Flow (our) demonstrates the fastest sampling time (ST) and competitive NLL with a smaller model size compared to other methods. All times are in milliseconds (ms) and parameters in millions (M).}
\resizebox{\columnwidth}{!}{
        \begin{tabular}{lrrrrr}
        \toprule
        \textbf{Method} & \textbf{ST (ms)}  & \textbf{FT (ms)}& \textbf{NLL} & \textbf{BPD} & \textbf{param} \\
        \midrule
        Emerging     & 332.7 $\pm 2.7$  & 121.0 $\pm 1.5$ & 630 & 1.12 & 0.16 \\
        FIncFlow     & 47.3 $\pm 2.3$  & 95.1 $\pm 2.5$ & 411 & 0.73 & 5.16 \\
        SNF          & 33.5 $\pm 2.2$ & 212.5 $\pm 7.3$ & 557   & 1.03 & 1.2\\
        Inverse-Flow & 12.2 $\pm 1.1$ & 77.9 $\pm 1.3$ & 350  & 0.62 & 0.6\\
        \bottomrule
        \end{tabular}
        }
    \label{tab:M_st_2_4}
\end{table}

\begin{table}[ht]
    \centering
    \caption{Performance comparison for MNIST with block size ($K = 16$) and number of blocks ($L = 2$). Inverse-Flow achieves the lowest sampling time (ST) while maintaining competitive performance in NLL and BPD with a smaller model size.}
    \resizebox{\columnwidth}{!}{
    \begin{tabular}{lrrrrr}
        \toprule
        \textbf{Method} & \textbf{ST} & \textbf{NLL} & \textbf{BPD} & \textbf{Param} & \textbf{Inverse}\\
        \midrule
        SNF          & 99 $\pm 2.1$ & 699 &  1.28  & 10.1  & approx \\
        FIncFlow     & 90 $\pm 2.2$ & 655 & 1.15  &  10.2 & exact \\
        MintNet      & 320$\pm 2.8$ & 630 & 0.98 & 125.9 &  approx \\
        Emerging     & 814$\pm 6.2$ & 640 & 1.09 & 11.4 &  exact \\
        Inverse-Flow & 52 $\pm 1.3$ & 710 & 1.31 & 1.6  & exact \\
        \bottomrule
        \end{tabular}
    }
    \label{tab:M_st_2_16}
\end{table}

\subsection{Modeling and Sample time for MNIST}
We compare sample time (ST) and number of parameters for small model architecture ($L = 2$, $K = 4$) on small image datasets, MNIST \citep{lecun1998gradient} with image size $1 \times28\times28$ in Table \ref{tab:M_st_2_4}. It may not be feasible to run huge models in production because of large computations. Therefore, it is interesting to study behavior of models when they are constrained in size. So, We compare Inverse-Flow with other Normalizing flow models with same number of flows per level ($K$), for $K = 4, 16$, and $L = 2$. In Table \ref{tab:M_st_2_4}, Inverse-Flow demonstrates the fastest ST of $12.2$, significantly outperforming others. This advantage is maintained in Table \ref{tab:M_st_2_16}, where Inverse-Flow achieves the second-best ST of $52\pm 1.3$, only behind SNF but with a much smaller parameter count. Inverse-Flow gives competitive forward time. Table \ref{tab:M_st_2_4} shows that Inverse-Flow has best forward time of $77.9 $, indicating efficient forward pass computations compared to other methods.

In Table \ref{tab:M_st_2_4}, Inverse-Flow achieves the lowest NLL ($350$) and BPD ($0.62$), suggesting superior density estimation and data compression capabilities for MNIST dataset with small model size. Inverse-Flow consistently maintains a low parameter count for all model sizes. Table \ref{tab:M_st_2_4} uses only $0.6$M parameters, which is significantly less than FInc Flow (5.16M) while achieving better performance. In Table \ref{tab:M_st_2_16}, Inverse-Flow has the smallest model size among all methods, demonstrating its efficiency. Inverse-Flow consistently shows strong performance across multiple metrics (ST, FT, NLL, BPD) while maintaining a compact model size. The following observations highlight Inverse-Flow's efficiency in sampling, density estimation, and parameter usage, making it a competitive method for generative modeling on the MNIST dataset.

For small linear flow architecture, our Inv\_Conv demonstrates the best sampling time of $19.7 \pm 1.2$, which is significantly faster than all other methods presented in Table \ref{tab:if_linear}. This indicates that Inv\_Conv offers superior efficiency in generating samples from a model, which is crucial for many practical applications of generative models. Inv\_Conv achieves the fastest forward time of $100$, outperforming all other methods. It has the smallest parameter count of 0.096 million, making it the most parameter-efficient approach. This combination of speed and compactness suggests that Inv\_Conv offers an excellent balance between computational efficiency and model size, which is valuable for deployment in resource-constrained environments or real-time applications.

\begin{table}[ht]
    \centering
    \caption{Runtime comparison of small planer models with $9$ layers with different invertible convolutional layers for MNIST. Inv\_Conv offers improved runtime efficiency with competitive NLL and fewer parameters compared to existing invertible convolutional layers.} 
    \resizebox{\columnwidth}{!}{
        \begin{tabular}{lrrrr}
        \toprule
        \textbf{Method} & NLL & \textbf{ST}  & \textbf{FT} & \textbf{Param} \\
        \midrule
        Exact Conv.       & 637.4 $\pm 0.2$ & 36.5 $\pm 4.1$ & 294 & 0.103 \\
        Exponential Conv. & 638.1 $\pm 1.0$ & 27.5 $\pm 0.4$ & 160 & 0.110 \\
        Emerging Conv.    & 645.7 $\pm 3.6$ & 26.1 $\pm 0.4$ & 143 & 0.103 \\
        SNF Conv.         & 638.6 $\pm 0.9$ & 61.3 $\pm 0.3$ & 255 & 0.364 \\
        Inv\_Conv (our)  & 645.3 $\pm 1.2$ & 19.7 $\pm 1.2$ & 100 & 0.096 \\
        \bottomrule
        \end{tabular}
        }
    \label{tab:if_linear}
\end{table}

\begin{table}[ht]
    \centering
    \caption{Performance comparison for CIFAR10 dataset with $L=2$ blocks and block size of $K=4$. Inverse-Flow achieves competitive BPD with significantly lower sampling time (ST) compared to existing methods while maintaining a small model size.}
    \resizebox{\columnwidth}{!}{
    \begin{tabular}{lrrrr}
    \toprule
    \textbf{Method} & \textbf{BPD}  & \textbf{ST} & \textbf{FT} & \textbf{Param} \\
    \midrule
    SNF              & 3.47  & 199.0 $\pm 2.2$ & 81.8 $\pm 3.6$  & 0.446\\
    Woodbury         & 3.55  & 2559.4 $\pm 10.5$ & 31.3 $\pm 1.5$ & 3.125 \\
    FIncFlow         & 3.52  & 47.3 $\pm 2.3$ & 125.5 $\pm 4.2$ & 0.589 \\
    Butterfly Flow   & 3.36  & 155 $\pm 4.6$ & 394.6 $\pm 3.4$ & 3.168 \\
    Inverse-Flow (our)     & 3.56  & 23.2 $\pm 1.3$ & 250.2 $\pm 2.9$ & 0.466 \\
    \bottomrule
    \end{tabular}
    }
    \label{tab:C_bpd_2_4}
\end{table}

\subsection{Modeling and Sample time for CIFAR10} 

In Table \ref{tab:C_bpd_2_4}, Inverse-Flow demonstrates the fastest sampling time of $23.2 \pm 1.3$, significantly outperforming other methods. This advantage is maintained in Table \ref{tab:bpd_2_16_C}, where Inverse-Flow achieves the second-best sampling time of $91.6 \pm 6.5$ among methods with exact inverse computation, only behind SNF which uses an approximate inverse. While not the fastest in forward time, Inverse-Flow shows balanced performance. In Table \ref{tab:C_bpd_2_4}, its forward time of 250.2 ± 2.9 is in the middle range. In Table \ref{tab:bpd_2_16_C}, its forward time of $722 \pm 7.0$ is competitive with other exact inverse methods.

\begin{table}[!ht]
    \centering
    \caption{Performance comparison for CIFAR dataset with block size ($K = 16$) and number of blocks ($L = 2$). Inverse-Flow achieves competitive BPD with significantly reduced sampling time (ST) compared to existing methods. SNF uses approx for inverse, and MintNet uses autoregressive functions. *time in seconds.}
    \resizebox{\columnwidth}{!}{
    \begin{tabular}{lrrrr}
    \toprule
    \textbf{Method} & \textbf{BPD} & \textbf{ST} & \textbf{FT} & \textbf{Param} \\
    \midrule
    SNF                 & 3.52 & 16.8 $\pm 2.7$ & 609 $\pm 5.4$ & 1.682  \\
    MintNet             & 3.51 & $25.0^*$ $\pm 1.5$ & 2458 $\pm 6.2$ & 12.466  \\
    \hline
    Woodbury            & 3.48 & 7654.4 $\pm 13.5$ & 119 $\pm 2.5$ & 12.49 \\
    MaCow               & 3.40 & 790.8 $\pm 4.3 $ & 1080 $\pm 6.6 $ & 2.68 \\
    CInC Flow           & 3.46 & 1710.0 $\pm 9.5$ & 615 $\pm 5.0$ & 2.62 \\
    Butterfly Flow      & 3.39 & 311.8 $\pm 4.0 $ & 1325 $\pm 7.5$ & 12.58\\
    FInc Flow           & 3.59 & 194.8 $\pm 2.5$ & 548 $\pm 6.2$ & 2.72 \\
    Inverse-Flow (our)        & 3.57 &  91.6 $\pm 6.5$ & 722 $\pm 7.0$ & 1.76 \\
    \bottomrule
    \end{tabular}
    }
\label{tab:bpd_2_16_C}
\end{table}

While not the best, Inverse-Flow maintains competitive BPD scores. In Table \ref{tab:C_bpd_2_4}, it achieves $3.56$ BPD, which is comparable to other methods. In Table \ref{tab:bpd_2_16_C}, its BPD of $3.57$ is close to the performance of other exact inverse methods. Inverse-Flow consistently maintains a low parameter count. In Table \ref{tab:C_bpd_2_4}, it uses only 0.466M parameters, which is among the lowest. In Table \ref{tab:bpd_2_16_C}, Inverse-Flow has the second-smallest model size ($1.76$M param) among methods with exact inverse computation, demonstrating its efficiency. Inverse-Flow demonstrates a good balance between sampling speed and BPD. Comparing Tables \ref{tab:C_bpd_2_4} and \ref{tab:bpd_2_16_C}, we can see that Inverse-Flow scales well when increasing the block size from 4 to 16. It maintains competitive performance across different model sizes and complexities. Table \ref{tab:bpd_2_16_C} highlights that Inverse-Flow provides exact inverse computation, a desirable property shared with several other methods like MaCow, CInC Flow, Butterfly Flow, and FInc Flow. For experiments on large image size, batch size,



%


\section{Conclusion}
We give a fast and efficient backpropagation algorithm for the inverse of convolution. Also, we proposed a flow-based model, Inverse-Flow, that leverages convolutions for efficient sampling and inverse of convolution for learning. Our key contributions include a fast backpropagation algorithm for the inverse of convolution, enabling efficient learning and sampling; a multi-scale architecture accelerating sampling in Normalizing flow models; a GPU implementation for high-performance computation; and extensive experiments demonstrating improved training and sampling timing. Inverse-Flow significantly reduces sampling time, making them competitive with other generative approaches. Our fast and efficient backpropagation opens new avenues for training more expressive and faster-Normalizing flow models. Inverse-Flow represents a substantial advancement in efficient, expressive generative modeling, addressing key computational challenges and expanding the practical applicability of flow-based models. This work contributes to the ongoing development of generative models and their real-world applications, positioning flow-based approaches as powerful tools in the machine-learning landscape.

\subsubsection*{Acknowledgments}
This research is funded by the iHub-IIITH PhD Fellowship 2023-24. We also extend our gratitude to IIIT-Hyderabad and ACM India for providing the International Travel Grant that enabled our participation in AISTATS'25. 
\bibliography{a_ref}
\bibliographystyle{abbrvnat} 

\onecolumn
\aistatstitle{Parallel Backpropagation for Inverse of a Convolution with Application to Normalizing Flows: Supplementary Materials}


We provide a comprehensive extension to the main paper, offering in-depth insights into the experimental setup, additional experimental results, and rigorous mathematical proofs. The Supplementary begins with experimental specifications Section \ref{sec:exp}, including information about model architecture, training parameters, and hardware used. In the next section \ref{sec:class}, we present an interesting application of inverse convolution layers in image classification, demonstrating high accuracy on the MNIST dataset with a remarkably small model. Section \ref{sec:proofs} presents thorough proofs of two theorems related to the backpropagation algorithm for the inverse of convolution layers. These proofs, presented with clear mathematical notation and step-by-step derivations, establish a theoretical foundation for computing input gradients and weight gradients in the context of inverse convolution operations.

\section{Input Padding and Kernel Masking}\label{sec:pad}

\begin{figure}[!ht]
    \centering
    \includegraphics[width=0.99\linewidth]{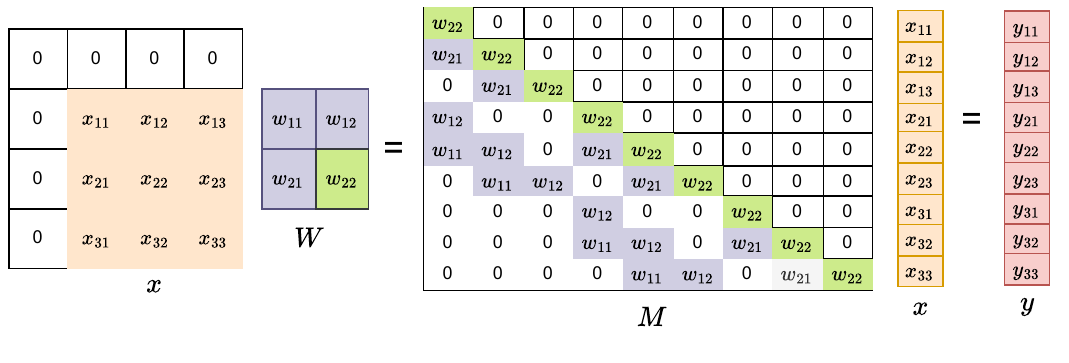}
    \caption{The convolution of a top-left padded image ($x_{m,m}$) with filter ($W$) involves a linear transformation of the input ($x$) via a lower triangular convolution matrix ($M_{m^2,m^2}$), where the diagonal elements correspond to the filter weights ($W_{k,k}$). Each row of $M$ computes a pixel, and these rows can be inverted in parallel using the steps outlined in the inversion algorithm by Kallappa et al. (2023).} 
    \label{fig:inv_conv_re}
\end{figure}

\section{Experimental Details}\label{sec:exp}
The architecture of SNF is the starting point for Inverse-Flow architecture and all our experiments. All models are trained using the Adam optimizer. We evaluate our Inverse-Flow model for density estimation (BPD, NLL), Sampling time (ST),  and Forward time (FT) with a batch size of $100$ for all experiments. For MNIST, we use an initial learning rate of $1e-3$, scheduled to decrease by one order of magnitude after 50 epochs for all datasets but CIFAR10, which is decreased every 25 epochs.
All the experiments were run on NVIDIA GeForce RTX 2080 Ti GPU. For MaCow, SNF, MaCow, and SNF, we use the official code released by the authors. Emerging was implemented in PyTorch by the authors of SNF, and we make use of that. We have implemented CInC Flow on PyTorch to get the results.

\section{Ablation Study} 
We conducted an ablation study on image sizes from $3 \times 16 \times 16$ to $3 \times 256 \times 256$ (the maximum input size for a single GPU). The results in the results section indicate that increasing image size minimally impacts model size and that both sampling and forward pass times increase linearly. Furthermore, we studied various kernel sizes (2x2, 3x3, 5x5, 7x7, 9x9, 11x11) and batch sizes (1, 2, 4, 8, 16, 32, 64, 128), which also showed a linear increase in sampling pass time, forward pass time, and required GPU memory. These details have been added to the supplement section for the final draft.
\paragraph{Input Size:} We performed ablation studies across varying input sizes (16x16 to 256x256). The results demonstrate that the proposed speedup in backpropagation is consistent across different input dimensions, with minimal variance in sampling time and predictable scaling in GPU memory usage. All tables attached below are for K=2, L=32 model architecture, and timing in milliseconds for Inverse-Flow Table-\ref{tab:if_ft_st_mem}, and FInC Flow Table-\ref{tab:ff_ft_st_mem}. Inverse-Flow achieves significantly lower sampling time across all input sizes than FInC Flow \ref{tab:ff_ft_st_mem} while maintaining similar GPU memory usage, though its forward time is slightly higher for larger input sizes.

\begin{table}[!ht]
    \centering
    \begin{tabular}{cccc}
        \toprule
        input size & Sampling Time (ST) & Forward Time (FT) & GPU memory (GB) \\
        \hline
        256x256 & 16.54 $\pm 0.21$ & 427 $\pm 20$ & 8.025\\
        128x128 & 16.22 $\pm 0.23$ & 342 $\pm 15$ & 3.822\\
          64x64 & 15.94 $\pm 0.14$ & 264 $\pm 11$ & 1.316\\
          32x32 & 15.53 $\pm 0.29$ & 224 $\pm 11$ & 0.375\\
          16x16 & 15.55 $\pm 0.12$ & 211 $\pm 10$ & 0.131\\
        \bottomrule
    \end{tabular}
    \caption{Inverse-Flow follows consistent sampling and forward times across varying input sizes with efficient GPU memory usage. Inverse-Flow: K = 2, L = 32, Sample size = 100, batch size = 100. All times in milliseconds}
    \label{tab:if_ft_st_mem}
\end{table}

As expected, the sampling and forward times linear increase with the input size, but our method maintains efficient memory usage even for larger images. For example, for a 256x256 input, the sampling time is 16.54 ms, the forward time is 427 ms, and the GPU memory usage is 8.025 GB. This shows that our approach scales reasonably well with input size while keeping the computational overhead manageable.

\begin{table}[!ht]
    \centering
    \begin{tabular}{cccc}
        \toprule
        input size & ST & FT & memory (GB) \\
        \hline
        256x256 & 23.08 $\pm 0.17$ & 285 $\pm 09$ & 8.016\\
        128x128 & 20.74 $\pm 0.15$ & 265 $\pm 15$ & 3.820\\
          64x64 & 17.99 $\pm 0.40$ & 243 $\pm 11$ & 1.314\\
          32x32 & 16.71 $\pm 0.12$ & 238 $\pm 11$ & 0.359\\
          16x16 & 17.21 $\pm 0.22$ & 255 $\pm 10$ & 0.129\\
        \bottomrule
    \end{tabular}
    \caption{FInC Flow: K = 2, L=32. Sample size =100, batch size = 100. All times in milliseconds}
    \label{tab:ff_ft_st_mem}
\end{table}

\paragraph{Benchmarking across batch sizes (1 to 128):} shows that GPU memory usage scales linearly with batch size, and forward timing performance remains efficient, supporting the adaptability of our method for real-world batch processing scenarios \ref{tab:if_ft_st_batch_size}.

\begin{table}[!ht]
    \centering
    \begin{tabular}{cccc}
        \toprule
        batch size & ST & FT & memory (GB) \\
        \hline
          128 & 19.65 $\pm 0.19 $  & 487 $ \pm 20 $ & 6.383\\
          64 & 18.99 $\pm 0.22 $  & 349 $ \pm 22 $ & 3.361\\
          32 & 18.83 $\pm 0.12 $  & 339 $ \pm 16 $ & 1.809\\
          16 & 18.48 $\pm 0.10$ & 335 $\pm 09$ & 1.037\\
          8 & 18.74 $\pm 0.15$ & 345 $\pm 11$ & 0.654\\
          4 & 18.04  $\pm 0.40$ & 333 $\pm 21$ & 0.457\\
          2 & 17.71 $\pm 0.12$ & 258 $\pm 11$ & 0.379\\
          1 & 16.22 $\pm 0.19$ & 235 $\pm 10$ & 0.328\\
        \bottomrule
    \end{tabular}
    \caption{Benchmarking across batch sizes (1 to 128) shows that GPU memory usage scales linearly with batch size while forward time performance remains efficient. Inverse Flow: K = 2, L=32. Sample size =100, batch size = 100. For 3x256x256, one NVIDIA GeForce GTX 1080 Ti
    GPU goes out of memory. All times in milliseconds}
    \label{tab:if_ft_st_batch_size}
\end{table}

\paragraph{Kernel Size:} Experiments with different convolution kernel sizes (2x2 to 11x11) reveal that while forward time increases with larger kernels, the sampling time remains relatively stable, validating the robustness of our approach to kernel size changes \ref{tab:if_ft_st_kernel_size}.

\begin{table}[!ht]
    \centering
    \begin{tabular}{ccccc}
        \toprule
        kernel size & ST & FT & Params (M) \\
        \hline
          11x11 & 17.87 $\pm 0.42 $  & 2428 $ \pm 28 $ & 6.068\\
          9x9 & 17.36 $\pm 0.22 $  & 1762 $ \pm 30 $ & 5.146\\
          7x7 & 19.20 $\pm 0.10$ & 1461 $\pm 23$ & 4.407\\
          5x5 & 20.79 $\pm 0.15$ & 934 $\pm 17$ & 3.856\\
          3x3 & 18.11 $\pm 0.20$ & 402 $\pm 08$ & 3.487\\
          2x2 & 17.90 $\pm 0.25$ & 364 $\pm 07$ & 3.372\\
        \bottomrule
    \end{tabular}
    \caption{Experiments with different kernel sizes (2x2 to 11x11) show that while forward time increases with larger kernels, sampling time remains stable, highlighting the robustness of our approach to kernel size variations. Inverse Flow model size and sampling time for different kernel sizes: Model Arch, K = 2, L=32. Sample size = 100, batch size = 100.  All times in milliseconds, M: millions.}
    \label{tab:if_ft_st_kernel_size}
\end{table}

\section{Image Classification using Inverse of Convolution Layers:}\label{sec:class}
\begin{figure}[!h]
    \centering
    \includegraphics[width=0.59\linewidth]{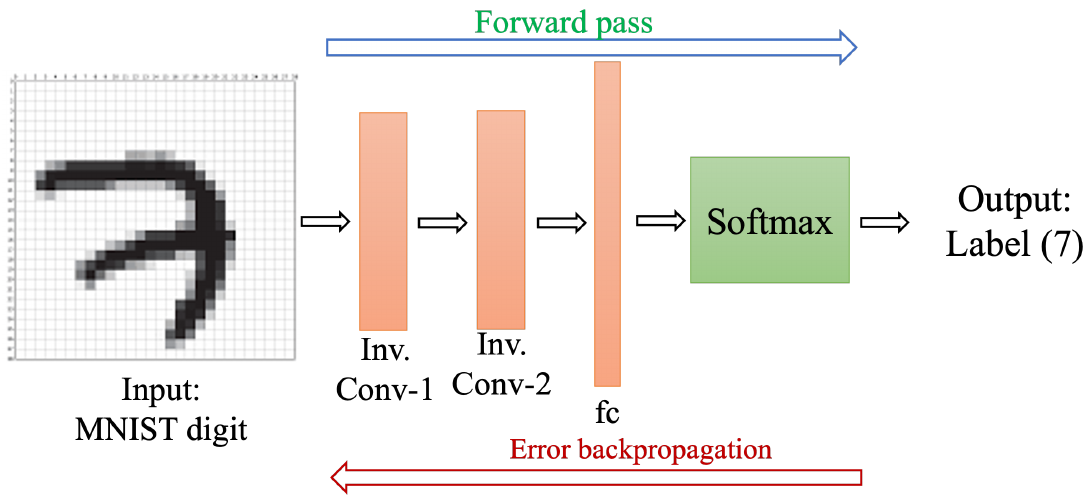}
    \caption{Overview of a small image classification model with two inverse of convolution ( $3 \times 3$ inv-con) layers with $97.6\%$ accuracy on MNIST dataset.}
    \label{fig:if_class}
\end{figure}

For MNIST digits image classification using the inverse of convolution (inv-conv) layers and proposed its backpropagation algorithm, we trained a two inv-conv layer and one fully connected layer model with 16 learnable parameters for inv-conv layers. See Figure \ref{fig:if_class}; this simple and small two inv-conv and one fully connected (FC) layers model gives $97.6\%$ classification accuracy after training for 50 epochs. 

\section{Formal Proofs}\label{sec:proofs}
In this section, we provide the proofs relating to the proposed backpropagation algorithm for the inverse of the convolution layer. First, we provide the following notation and the equation for the gradients.
\paragraph{Notation:} We will follow the notation used in the main paper. \\ We will denote input to the inverse of convolution (inv-conv) by $y \in \mathbb R^{m^2}$ and output to be $x \in \mathbb R^{m^2}$. We will be indexing $x,y$ using $p = (p_1, p_2) \in \{ 1, \cdots, n\} \times \{ 1, \cdots, n\}$. We define 
$$\Delta(p) = \{ (i,j):  0 \leq p_1 - i, p_2 -j < k  \} \setminus \{ p \}.$$
$\Delta(p)$ informally is  set of all pixels except $p$ which depend on $p$, when convolution is applied with top, left padding. We also define a partial ordering $\leq$ on pixels as follows
$$ p \leq q \quad \Leftrightarrow \quad p_1 \leq q_1 \text{ and } p_2 \leq q_2.$$
The kernel of $k\times k$ convolution is given by matrix $W \in \mathbb R^{k \times k}$. For the backpropagation algorithm for inv-conv, the input is 
$$x \in \mathbb R^{m^2} \text{ and } \frac{\partial L}{\partial x} \in \mathbb R^{m^2},$$ 
where $L$ is the loss function. We can compute $y$ on $O(mk^2)$ time using the parallel forward pass algorithm Aaditya et. al. The output of backpropagation algorithm is $$\frac{\partial L}{\partial y} \in \mathbb R^{m^2} \text{ and } \frac{\partial L}{\partial W} \in \mathbb R^{k^2}$$
which we call input and weight gradient, respectively. We provide the algorithm for computing these in the next 2 subsections.
\subsection{Theorem 1}
\paragraph{Computing Input Gradients}
Since $y$ is input to inv-conv and $x$ is output, $y = \text{conv}_W(x)$ and we get the following $m^2$ equations by definition of the convolution operation. 
\begin{equation}
\label{eqn:inv_conv}
  y_p = xp + \sum_{q \in \Delta(p)} W_{(k,k) - p + q} \cdot  x_q 
\end{equation}
Using the chain rule of differentiation, we get that
\begin{equation}
    \frac{\partial L}{ \partial y_p} = \sum_{q} \frac{\partial L}{ \partial x_q} \times \frac{\partial x_q}{ \partial y_p}.
\end{equation}
Hence if we find $\frac{\partial x_q}{ \partial y_p}$ for every pixels $p,q$, we can compute $ \frac{\partial L}{ \partial y_p}$ for every pixel $p$.

\paragraph{Theorem 1:} Input $y$ gradients

\begin{equation}
\frac{\partial x_q}{ \partial y_p} = 
\begin{cases}
    1 & \text{ if } p = 1\\
    0 & \text{ if } q \not \leq p\\
    - \sum_{r \in \Delta(p)} W_{(k,k) - r} \frac{\partial x_{p-r'}}{\partial y_p} & \text{ otherwise. }
\end{cases}
\end{equation}

\begin{proof} We will prove Theorem \ref{the:dx_dy} by induction on the partial ordering of pixels.

\textbf{Base Case:} For $p = (1,1)$, which is the smallest element in our partial ordering:

From Equation (\ref{eqn:inv_conv}), we have: $y_{(1,1)} = x_{(1,1)}$ . This implies: $\frac{\partial x_{(1,1)}}{\partial y_{(1,1)}} = 1$ and for any $q \neq (1,1)$: $\frac{\partial x_q}{\partial y_{(1,1)}} = 0$. This satisfies the theorem for the base case.

\textbf{Inductive Step:} Assume the theorem holds for all pixels less than $p$ in the partial ordering. We will prove it holds for $p$.

\begin{enumerate}
    \item For $q \not\leq p$, $x_q$ does not depend on $y_p$ due to the structure of the convolution operation. Therefore, $\frac{\partial x_q}{\partial y_p} = 0$.

    \item For $q \leq p$, we differentiate Equation (\ref{eqn:inv_conv}) with respect to $y_p$:

    \[\frac{\partial x_p}{\partial y_p} = \frac{\partial x_p}{\partial y_p} + \sum_{r \in \Delta(p)} W_{(k,k) - p + r} \cdot \frac{\partial x_r}{\partial y_p}\]

    \begin{equation} \label{eqn_one}
        1 = \frac{\partial x_p}{\partial y_p} + \sum_{r \in \Delta(p)} W_{(k,k) - p + r} \cdot \frac{\partial x_r}{\partial y_p}
    \end{equation}

    Rearranging \ref{eqn_one}:

    \begin{equation}
    \frac{\partial x_p}{\partial y_p} = 1 - \sum_{r \in \Delta(p)} W_{(k,k) - p + r} \cdot \frac{\partial x_r}{\partial y_p}
    \end{equation}

    This is equivalent to the third case in the theorem, with $q = p$.

    \item For $q < p$, we can write:

    \[x_q = y_q - \sum_{r \in \Delta(q)} W_{(k,k) - q + r} \cdot x_r\]

    Differentiating with respect to $y_p$:

    \[\frac{\partial x_q}{\partial y_p} = \frac{\partial y_q}{\partial y_p} - \sum_{r \in \Delta(q)} W_{(k,k) - q + r} \cdot \frac{\partial x_r}{\partial y_p}\]

    Since $q < p$, $\frac{\partial y_q}{\partial y_p} = 0$. Therefore:

    \begin{equation}
    \frac{\partial x_q}{\partial y_p} = - \sum_{r \in \Delta(q)} W_{(k,k) - q + r} \cdot \frac{\partial x_r}{\partial y_p}
    \end{equation}

    This is equivalent to the third case in the theorem.
\end{enumerate}
Thus, by induction, the theorem holds for all pixels $p$.
\end{proof}
\subsection{Theorem 2}

\paragraph{Computing Weight Gradients}
From Equation \ref{eqn:inv_conv}, we can say computing the gradient of loss $L$ with respect to weights $W$ involves two key factors. Direct influence: how a specific weight $W_{}$ in convolution kernel directly affects output $x$ pixels, and Recursive Influence: how neighboring pixels, weighted by the kernel, indirectly influence output $x$ during the inverse of convolution operation. Similarly, to compute  gradient of  loss $L$ w.r.t filter weights \( W \), we apply  chain rule:

\begin{equation}\label{eqn:dl_dw}
    \frac{\partial L}{\partial W} = \frac{\partial L}{ \partial x} \times \frac{\partial x}{\partial W}
\end{equation}

where: \( \frac{\partial L}{\partial x} \) is  gradient of  loss with respect to  output \( x \) and inverse of convolution operation is applied between \( \frac{\partial L}{\partial x} \) and  output \( x \). Computing the gradient of loss $L$ with respect to convolution filter weights $W$ is important in backpropagation when updating the convolution kernel during training.
Similarly, $\partial L/ \partial W$ can be calculated as Equation \ref{eqn:dl_dw} and $\partial x/\partial W$ can be calculated as (Equation \ref{eqn:dl_dw}) for each $k_{i, j}$ parameter by differentiating Equation \ref{eqn:inv_conv} w.r.t $W$:

\begin{equation}
    \label{eqn-dw}
    \frac{\partial L}{ \partial W_a} =  \sum {\frac{\partial L}{\partial x_q} \times \frac{\partial x_q}{\partial W_a}}
\end{equation}

Equation \ref{eqn-dw} states that to compute the gradient of the loss with respect to each weight $W_a$, we need to:
\begin{itemize}
    \item Compute how loss $L$ changes with respect to each output pixel $x_q$ (denoted by $\frac{\partial L}{ \partial x_q}$).
    \item Multiply this by gradient of each output pixel $x_q$ with respect to weight $W_a$ (denoted by $\frac{\partial x_q}{ \partial W_a}$)
\end{itemize}

We then sum over all output pixels $x_q$.

\paragraph{Theorem 2:} Weights $W$ gradients

\begin{equation} 
\frac{\partial x_q}{\partial W_a} = 
\begin{cases}
    0 & \text{if } q \leq a \\
    -\sum_{q' \in \Delta q(a)} W_{q' - a} \times \frac{\partial x_{q-q'}}{\partial W_a} - x_{q-a} & \text{if } q > a
\end{cases}
\end{equation}

\begin{proof}
We will prove Theorem \ref{eq:dx_dw} by induction on the partial ordering of pixels.

\textbf{Base Case:}

For $q \leq a$, we have $\frac{\partial x_q}{\partial W_a} = 0$. 

This is because, in the inverse of convolution operation, $x_q$ does not directly depend on $W_a$. The weight $W_a$ only affects pixels that come after $q$ in the partial ordering.


\textbf{Inductive Step:} Assume the theorem holds for all pixels less than $q$ in the partial ordering. We will prove it holds for $q > a$.

From Equation (\ref{eqn:inv_conv}), we have:

\begin{equation}\label{eqn:yq_xq}
y_q = x_q + \sum_{r \in \Delta(q)} W_{(k,k) - q + r} \cdot x_r
\end{equation}

Rearranging this equation \ref{eqn:yq_xq}:

\begin{equation}\label{eqn:xq_yq}
x_q = y_q - \sum_{r \in \Delta(q)} W_{(k,k) - q + r} \cdot x_r
\end{equation}

Now, let's differentiate both sides of \ref{eqn:xq_yq}with respect to $W_a$:
\begin{equation}\label{eqn:xq_wa}
\frac{\partial x_q}{\partial W_a} = \frac{\partial y_q}{\partial W_a} - \sum_{r \in \Delta(q)} \left(\frac{\partial W_{(k,k) - q + r}}{\partial W_a} \cdot x_r + W_{(k,k) - q + r} \cdot \frac{\partial x_r}{\partial W_a}\right)
\end{equation}

Note that $\frac{\partial y_q}{\partial W_a} = 0$ because $y$ is the input to the inverse convolution and doesn't depend on $W$.

Also, $\frac{\partial W_{(k,k) - q + r}}{\partial W_a} = 1$ if $(k,k) - q + r = a$, and 0 otherwise.

Let $\Delta q(a) = \{r \in \Delta(q) : (k,k) - q + r = a\}$. Then we can rewrite the equation \ref{eqn:xq_wa} as \ref{eqn:xq_wa_next}:
\begin{equation}\label{eqn:xq_wa_next}
\frac{\partial x_q}{\partial W_a} = - \sum_{r \in \Delta q(a)} x_r - \sum_{r \in \Delta(q)} W_{(k,k) - q + r} \cdot \frac{\partial x_r}{\partial W_a}
\end{equation}

The first sum simplifies to $-x_{q-a}$ because $r = q - (k,k) + a$ for $r \in \Delta q(a)$.

In the second sum, we can use the inductive hypothesis for $\frac{\partial x_r}{\partial W_a}$ because $r < q$.

Therefore:
\begin{equation}\label{eqn:last}
\frac{\partial x_q}{\partial W_a} = -x_{q-a} - \sum_{r \in \Delta(q)} W_{(k,k) - q + r} \cdot \frac{\partial x_r}{\partial W_a}
\end{equation}

The right side of \ref{eqn:last} is equivalent to the second case in the theorem.

Thus, by induction, the theorem holds for all pixels $q$.
\end{proof}
\vfill

\end{document}